\pdfoutput=1

\documentclass[11pt]{article}

\usepackage[]{acl}

\usepackage{multicol}

\usepackage{graphicx}
\usepackage{caption}

\usepackage{times}
\usepackage{latexsym}
\usepackage{tabularx}
\usepackage{url}
\usepackage{amsmath}
\usepackage[T1]{fontenc}

\usepackage[utf8]{inputenc}

\usepackage{microtype}

\usepackage{inconsolata}

\newcommand{\datasetname}[0]{GuyLingo}

\title{\datasetname: The Republic of Guyana Creole Corpora}

\author{
  Christopher Clarke$^*$$^{1,2}$\hspace{10pt} Roland Daynauth$^*$$^{1,2}$\hspace{10pt} Charlene Wilkinson$^2$\hspace{10pt} \\ \textbf{Hubert Devonish\hspace{10pt} Jason Mars$^1$\hspace{10pt}}   \vspace{0.3cm}\\
    \text{$^1$University of Michigan, Ann Arbor, MI}\\
    \text{$^2$University of Guyana, Georgetown, Guyana}\\
    \text{\{csclarke, daynauth, profmars\}@umich.edu} \\
    \text{\{christopher.clarke, roland.daynauth, charlene.wilkinson\}@uog.edu.gy} \\
    \text{hubertsldevonish@gmail.com}
}

\begin{document}
\maketitle
\begin{abstract}
While major languages often enjoy substantial attention and resources, the linguistic diversity across the globe encompasses a multitude of smaller, indigenous, and regional languages that lack the same level of computational support. One such region is the Caribbean. While commonly labeled as "English speaking", the ex-British Caribbean region consists of a myriad of Creole languages thriving alongside English. In this paper, we present \textbf{\datasetname}: a comprehensive corpus designed for advancing NLP research in the domain of Creolese (Guyanese English-lexicon Creole), the most widely spoken language in the culturally rich nation of Guyana. We first outline our framework for gathering and digitizing this diverse corpus, inclusive of colloquial expressions, idioms, and regional variations in a low-resource language. We then demonstrate the challenges of training and evaluating NLP models for machine translation in Creole. Lastly, we discuss the unique opportunities presented by recent NLP advancements for accelerating the formal adoption of Creole languages as official languages in the Caribbean\footnote{\url{https://github.com/ChrisIsKing/Caribbean-Creole-Languages-Translation}}.
\end{abstract}

\begingroup\def\thefootnote{*}\footnotetext{Equal contribution.}\endgroup

\begin{figure}
    \centering
    \includegraphics[width=0.75\columnwidth]{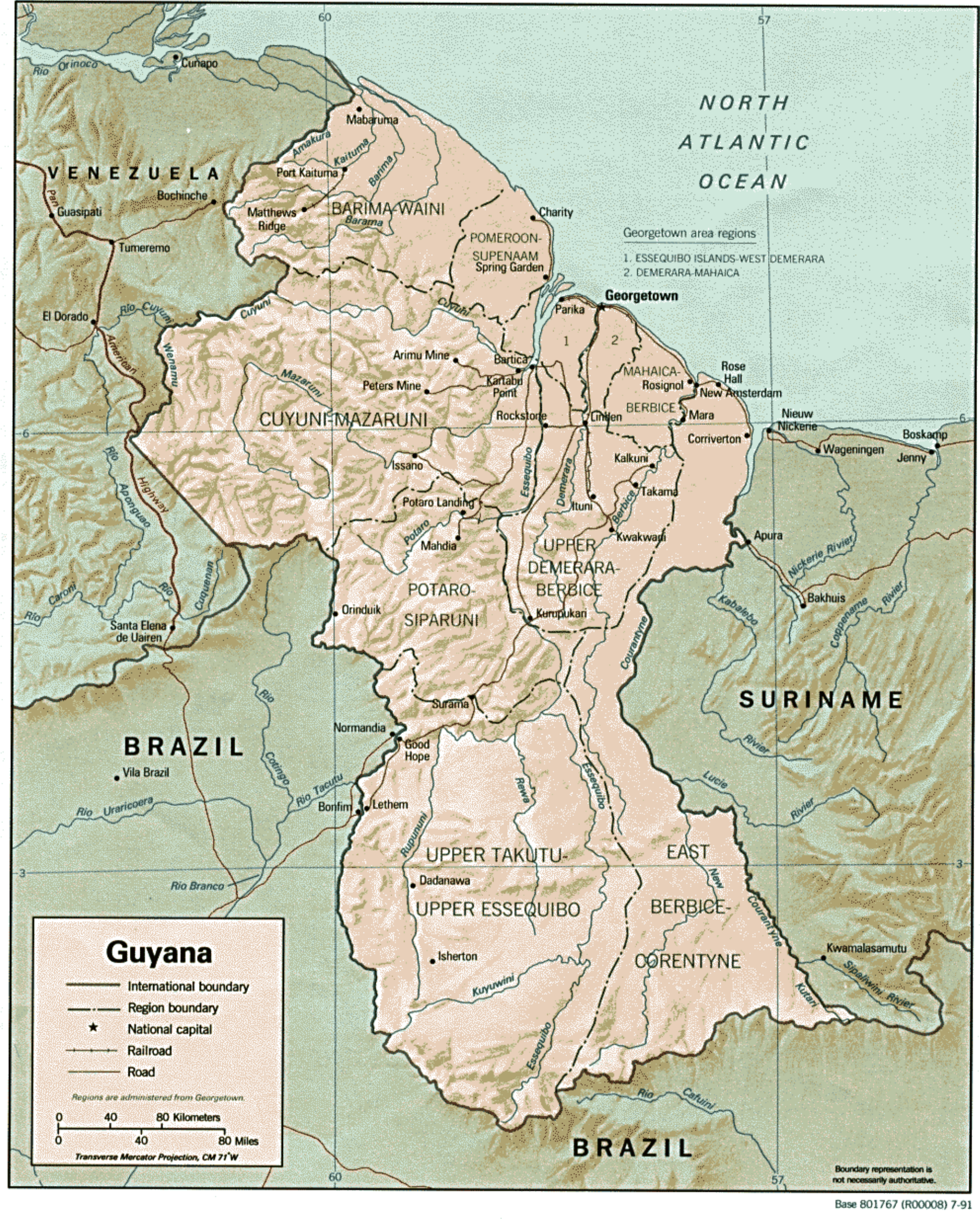} 
    \caption{Map of Guyana and its neighboring territories}
    \label{fig:map}
    \captionsetup{font=small}
\end{figure}

\section{Introduction}
Major languages such as English and Chinese frequently receive considerable attention and resources due to their global prominence and economic influence \cite{lent-etal-2021-language, lent-etal-2022-ancestor}. The extensive focus on these major languages in natural language processing (NLP) has resulted in the development of sophisticated models, extensive datasets, and digital applications consumed by millions of users today. However, despite this global prominence, the linguistic landscape of the globe extends far beyond these dominant languages, encompassing a plethora of smaller, indigenous, and regional languages that play crucial roles in the cultural heritage and communication of their respective communities \cite{lent-etal-2022-creole, hershcovich2022challenges}. The countries of the Commonwealth (ex-British) Caribbean Community represent an example cluster of such countries.

\begin{table*}[]
    \centering
    \resizebox{\textwidth}{!}{%
    \begin{tabular}{p{10cm}|p{10cm}}
        \hline
        \textbf{Creole} & \textbf{English} \\
        \hline
       ``It luk laik nof ting cheenj op,” Seera se. ``Somtaim mi doz get fraikn.” & “So many things feel like they have changed,” said Sara. “I get scared about it sometimes.” \\
        \hline
        When me lef' han' 'cratch me, money a-come & When my left hand itches, money is coming. \\
        \hline
        Di leedii prapa nais & The lady is very pretty \\
        \hline
    \end{tabular}}
    \caption{Example Guyanese Creole from \datasetname{} and its English Translation}
    \label{tab:creole_english_type}
\end{table*}

Within the diverse linguistic tapestry of the Caribbean Community, a rich array of languages thrives, reflecting the historical, cultural, and ethnic diversity of the region \cite{rickford1987dimensions, holbrook2001guyanese}. While English is commonly used as the official language in many Commonwealth Caribbean states, the linguistic heritage goes beyond just English, encompassing a variety of Creole languages, indigenous languages, and influences from African, Indigenous, European, and Asian languages \cite{DevonishThompson2013}.


Creole languages of the Caribbean emerged out of the language contact between Europeans and Africans arising from colonialism and plantation slavery. These languages, such as Jamaican, otherwise referred to as Jamaican Creole or Jamaican Patois/Patwa \cite{armstrong-etal-2022-jampatoisnli}, Trinidadian Creole \cite{apics}, and Haitian Creole \cite{hewavitharana-etal-2011-cmu}, have evolved as vibrant means of communication, showing language features originating in West African languages as well as the languages of the colonizing Europeans. \cite{hagemeijer-etal-2014-gulf}.

Despite its prominence as the mother tongue of the majority of the over 700,000 inhabitants of the Republic of Guyana,
Creolese (Guyanese English-lexicon Creole) is a low-status vernacular language that takes second place to the sole official language, English. This is typical of local vernaculars in post-colonial situations like Guyana \cite{hershcovich2022challenges}. English has been traditionally the only language in which Guyanese children are taught to read and write in school. Written resources in Creolese are limited, making it a low-resource language within the field of Natural Language Processing (NLP).




In this work, we introduce \datasetname{}, a corpus for Creolese curated for advancing NLP research and development in Creole. Using this resource, we explore the task of machine translation between English and Creolese. To aid in this process we design and implement the Guyanese Creole Translation tool\footnote{\url{https://translation.csclarke.com}}, a web-based GPT-powered machine translation tool. Lastly, we briefly discuss the insights gained from developing Guylingo alongside considerations for accelerating the formal adoption of Creole languages as official languages in the Caribbean.



\section{\datasetname{} Corpus}

This section describes the curation of \datasetname{}, a corpus of Creolese, the primary spoken language of Guyana. The creation of this corpus aims to address the scarcity of resources and attention devoted to indigenous and regional languages within the NLP community.

\subsection{Data Collection}
The compilation of \datasetname{} requires the collecting and digitizing of a series of linguistic resources. These sources should ideally encompass a spectrum of Creolese expressions, idiomatic phrases, and regional variations. To ensure inclusivity and authenticity, we employ a multi-pronged approach.

\subsubsection{Expert Collaboration}
In collaboration with the University of Guyana, Guyanese Languages Unit, a collection of original Guyanese Creole sources was curated, digitized, and manually transcribed by a team of researchers. Examples of this include \citet{peirs1902proverbs} a book of Guyanese proverbs, containing over 1k culturally rich proverbs from early British Guiana times still used today, and \citet{unicef-my-hero-you} a COVID-19 children's book transcribed by Creolese experts for primary education students. In addition, our team of native Creole experts manually construct a corpus of high-quality common Guyanese Creole sayings and terms. Table \ref{tab:data_source} shows a full breakdown of all information sources.

\subsubsection{Online Resources}
Whilst some of the aforementioned sources use the consistent phonemic Cave-GLU standard writing system \cite{cave-glu} for creole, others do not. This is particularly true for the many web-based sources such as language forums, blogs, educational platforms, etc., that contain small excerpts of colloquialisms, everyday conversations, and idiomatic expressions prevalent in the Guyanese Creole. These sources were scraped, cleaned, verified, and added to \datasetname{} as shown in \ref{tab:data_source}.

\begin{table*}[t]
    \centering
    \small
    \begin{tabularx}{\textwidth}{X >{\hsize=.4\hsize}X c c}
        \hline
        \textbf{Sources} & \textbf{Type} & \textbf{\# Sentences} & \textbf{Vocab Size} \\
        \hline
        Guyanese-Creole-English Vocabulary-Basic words. \cite{polyglotclub-guyanese-creole} & Corpus & 20 & 71 \\ 
        \hline
        Guyanese Creole. \cite{wiki-guyanese-creole} & Article & 6 & 28 \\ 
        \hline
        Gender and Pronominal Variation in an Indo-Guyanese Creole-Speaking \cite{sidnell1999gender} & Journal Article & 21 & 82 \\ 
        \hline
        Review of Guyanese Creole English \cite{GuyaneseCreoleEnglish} & Presentation & 28 & 96\\ 
        \hline
        Guyanese Creole Survey Report. \cite{holbrook2001guyanese} & Language Survey & 8 & 45\\ 
        \hline
        APiCS Online -Structure dataset. \cite{apics} & Report & 344 & 351 \\
        \hline 
        Creolese. \cite{DevonishThompson2013} & Journal Article & 69 & 112\\
        \hline 
        Habitual and Imperfective in Guyanese Creole. \cite{sidnell2002habitual} & Journal Article & 60 & 103\\ 
        \hline
        Tense and aspect in Guyanese Creole: A syntactic, semantic and pragmatic analysis \cite{gibson1982tense} & PhD Thesis & 231 & 374\\
        \hline 
        Two areas of Guyanese Grammar \cite{GuyaneseGrammar} & Article & 14 & 26\\
        \hline 
        Me Na Able: Creolese 101 \cite{LettersFromGuyana} & Blog & 9 & 25\\
        \hline 
        Travel Phrases - Guyanese Creole \cite{TravelPhrasesGuyanese} & Blog & 4 & 9\\
        \hline 
        My Hero is you \cite{unicef-my-hero-you} & Educational & 322 & 831 \\
        \hline
        The Proverbs of British Guiana \cite{peirs1902proverbs} & Book & 905 & 2054\\
        \hline
        Common Guyanese Creole Sayings (Manually created by experts)  & Corpus & 332 & 712\\
        \hline
        \textbf{Total} & & \textbf{2373} & \textbf{4177}\\
        \hline
    \end{tabularx}
    \caption{Compilation of Guyanese Creole Language Resources: Sources, Type, Sentences, and Vocabulary Size}
    \label{tab:data_source}
\end{table*}

\subsection{Dataset Characteristics}
\datasetname{} encapsulates a diverse array of linguistic data, including but not limited to:
\begin{itemize}
\setlength\itemsep{0.3em}
    \item Conversational dialogues
    \item Idiomatic expressions and phrases
    \item Proverbs and folklore
    \item Regional variations and dialectical nuances
\end{itemize}

In total, \datasetname{} consists of 2373 Guyanese Creole sentences with a vocabulary size of 4177 unique Creole words.

\section{\datasetname{} for Machine Translation}
To investigate the utility of \datasetname{}, we conduct experiments on the task of machine translation assessing the ability of NLP models to facilitate English$\leftarrow\rightarrow$Guyanese Creole translation. As such to enable the training and evaluation of these models \datasetname{} was further expanded to include English Creole translation pairs. Of the 2373 sentences, the Common Guyanese Creole Sayings corpus (expertly curated) was manually transcribed into English. The remaining Creole sources were extracted alongside their English translations. In addition, 339 common Creole terms from \citep{peirs1902proverbs} alongside their English pairs were extracted and verified. Using these initial translation pairs, the Guyanese Creole Translation Tool was built to allow the initial translation of remaining sentences in \datasetname{}. Of the remaining Creole sources, the largest source of Creolese data from \citet{peirs1902proverbs} contained no English translations for the proverbs. As such, using the translation tool these proverbs were machine translated. These machine translations were reviewed and edited for lexical correspondence but not semantic meaning given the complexity of translating the contextual meaning of cultural proverbs.

\begin{figure}[!t]
    \centering
    \includegraphics[width=0.8\columnwidth]{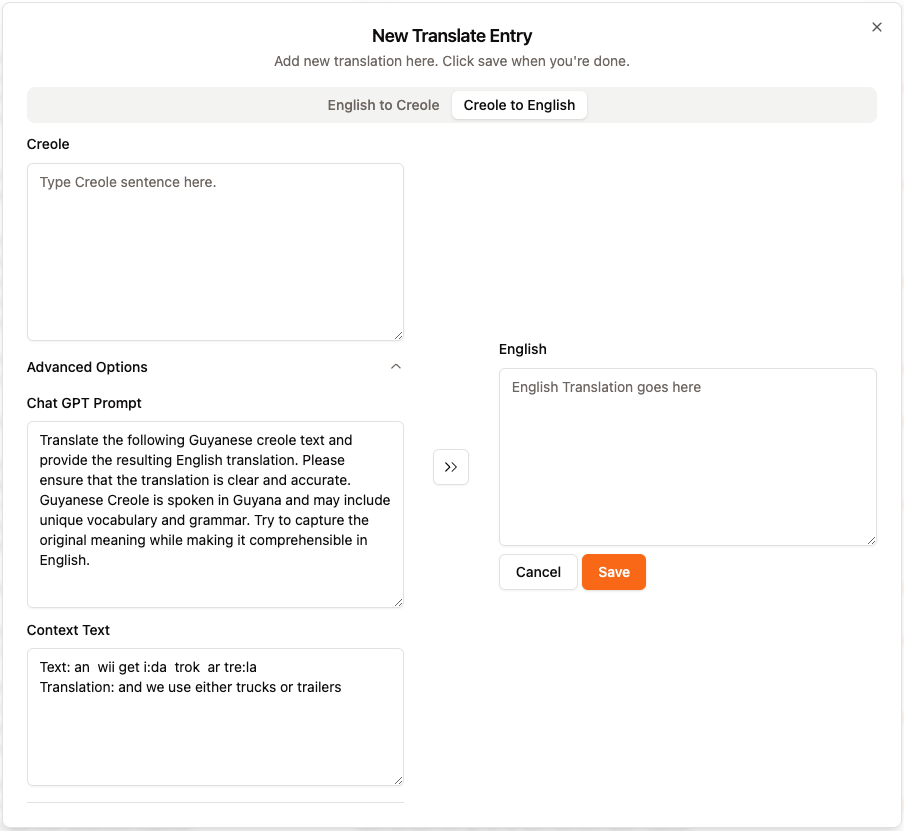} 
    \caption{User Interface of Guyanese Creole Translation Tool. This tool allows experts to rapidly and iteratively create translation pairs using GPT-4 \cite{openai2023gpt4} as a generator.}
    \label{fig:tool}
    \captionsetup{font=small}
\end{figure}

\subsection{Guyanese Creole Translation Tool}
The Guyanese Creole Translation tool, as shown in figure \ref{fig:tool}, is a web-based application built using Django+React to facilitate easy storing, editing, and iterative testing of English Creole translations. The UI allows Creolese experts to easily enter text in English or Creole and get a sample translation. We utilize GPT-4 \cite{openai2023gpt4} to automatically perform these translations. The advanced prompt includes a subset of example verified translations from \datasetname{} as in-context examples for generation. Once prompted, the user can modify the generated output before saving it to the database. Users also have the option to modify the advanced prompts as well as provide more seed examples for greater control over the translation process. For instance, users can provide a Guyanese proverb and instruct GPT4 to consider the nuances of the Guyanese culture while translating the text. As mentioned previously, a subset of \datasetname{}'s English pairs was generated using this tool. 

\paragraph{Translation Dataset Statistics} In total, our translation dataset consists of 1969 total translation pairs. For training and evaluation, we use the 302 manually curated translation pairs for testing and the remaining \datasetname{} translation pairs for model training. The manually curated translation pairs are all written in the Cave-GLU standard phonemic system for Creoles \cite{cave-glu}.

\subsection{Experiment Setup}

\paragraph{Training and Models} \label{train}
We consider the models of T5 \cite{raffel2023exploring}, BART \cite{lewis-etal-2020-bart} and Pegasus \cite{zhang2020pegasus} for their demonstrated performance on several machine translation tasks. All models were implemented with PyTorch and Hugging Face Transformers. We train all models with AdamW \citep{adamw} and a weight decay of 0.01. We use a learning rate of 2e-5, batch size of 4, and a linear learning rate warmup over the first 10\% steps with a cosine schedule. We pre-process the data and train all models with varying random seeds over multiple runs for 10 epochs. Approximately 200 GPU hours were required to train all hyperparameter variations across all tasks. Additionally, we evaluate the performance of GPT-4 using in-context learning on \datasetname{}.

\paragraph{Evaluation} For evaluation metrics, we adopted common automatic evaluation methods used for language generation based on n-gram overlap: BLEU \cite{Papineni02bleu:a}, ROUGE \cite{lin-2004-rouge}, METEOR \cite{banarjee2005} and CHRF \cite{popovic-2015-chrf}.

\begin{table}[]
    \centering
    \resizebox{\columnwidth}{!}{%
    \begin{tabular}{c|c|c|c|c|c|c}
        \hline
        \textbf{Model} & \textbf{Bleu} & \textbf{Rouge1} & \textbf{Rouge2} & \textbf{RougeL} & \textbf{Meteor} & \textbf{CHRF} \\
        \hline

GPT-4 (Zero-shot)	& 1.35	& 17.22	& 2.4	& 17.0	& 12.55 &	21.68 \\
        \hline

GPT-4 (Few-shot)& 1.64	& 20.6	& 3.42	& 20.2	& 14.56	& 22.32 \\
\hline
        
        T5-Large & 09.74 & 37.44 & 13.74 & 36.63 & 28.19 & 30.09 \\
        \hline
        Bart-Large & \textbf{12.11} & \textbf{40.56} & \textbf{18.47} & \textbf{39.64} & \textbf{32.77} & \textbf{33.21}\\
        \hline
        Bart-Base & 10.17 & 37.49 & 16.08 & 36.59 & 29.54 & 29.47\\
        \hline
        Pegasus-Large & 02.67 & 24.15 & 05.30 & 23.16 & 16.38 & 19.69 \\
        \hline
    \end{tabular}}
    \caption{Performance of MT Models on English-Creole Translation}
    \label{tab:evaluation_metrics}
\end{table}

\begin{table}[]
    \centering
    \resizebox{\columnwidth}{!}{%
    \begin{tabular}{c|c|c|c|c|c | c}
        \hline
        \textbf{Model} & \textbf{Bleu} & \textbf{Rouge1} & \textbf{Rouge2} & \textbf{RougeL} & \textbf{Meteor} & \textbf{CHRF}\\
        \hline

GPT-4 (Zero-shot) &	29.8	&60.4	&38.25	&58.96	&52.94	&51.33\\
\hline
GPT-4 (Few-shot) &	\textbf{30.24}	& \textbf{60.6}	& \textbf{39.31}	& \textbf{58.99}	& \textbf{54.14}	& \textbf{51.61} \\
\hline
        T5-Large & 19.70 & 47.71 & 26.99 & 46.47 & 42.45 & 39.89\\
        \hline
        Bart-Large & 17.70 & 45.74 & 24.41 & 39.75 & 32.77 & 37.23\\
        \hline
        Bart-Base & 14.20 & 41.68 & 20.04 & 40.40 & 35.95 & 34.33 \\
        \hline
        Pegasus-Large & 6.10 & 28.96 & 09.88 & 27.91 & 22.53 & 23.72\\
        \hline
    \end{tabular}}
    \caption{Performance of MT Models on Creole-English Translation}
    \label{tab:evaluation_metrics2}
\end{table}

\section{Results}
Table \ref{tab:evaluation_metrics} and \ref{tab:evaluation_metrics2} summarize our evaluation results on automated metrics. For en-creole translation, our results show that the Bart-Large model achieves the best performance amongst all models with a BLEU score of 12.11, ROUGE-1 score of 40.56, ROUGE-2 score of 18.47, ROUGE-L score of 39.64, METEOR score of 32.77 and a CHRF score of 33.21 outperforming other fine-tuned models such as T5 and large language models such as GPT-4 both in zero and few-shot prompting settings. The performance of en-creole translation is due to a couple of factors: 1) \textit{Incoherent English to Creole mapping}: Many unique words/phrases found in Creole do not contain English equivalents. 2) \textit{ Writing System of Testing set:} The Creolese Cave-GLU writing system \cite{cave-glu} used by the manually curated testing set is phonemic whereby a particular sound is always represented by one letter/letter combination. \datasetname{}'s training data, however, contains samples where multiple letter/letter combinations represent one sound. Thus, the chance of an error is much higher for en-creole translations.

For creole-eng translation, GPT-4 (Few-shot) using a subset of \datasetname{} training set as in-context learning examples delivers the best performance with a BLEU score of 30.24. a ROUGE-1 score of 60.6, a ROUGE-2 score of 39.31, a ROUGE-L score of 58.99, a METEOR score of 54.14, and a CHRF score of 51.61. This result highlights the power of GPT-4's large and extensive training on a diverse and extensive range of text data in addition to its ability to quickly adapt to new tasks or language pairs with only a few examples.




\section{Discussion}
In this section, we briefly discuss the unique opportunity presented by recent NLP advancements for accelerating the formal adoption of Creole languages in the Caribbean.


\subsection{AI-Driven Applications for Native Languages}
\begin{figure}
    \centering
    \includegraphics[width=0.8\columnwidth]{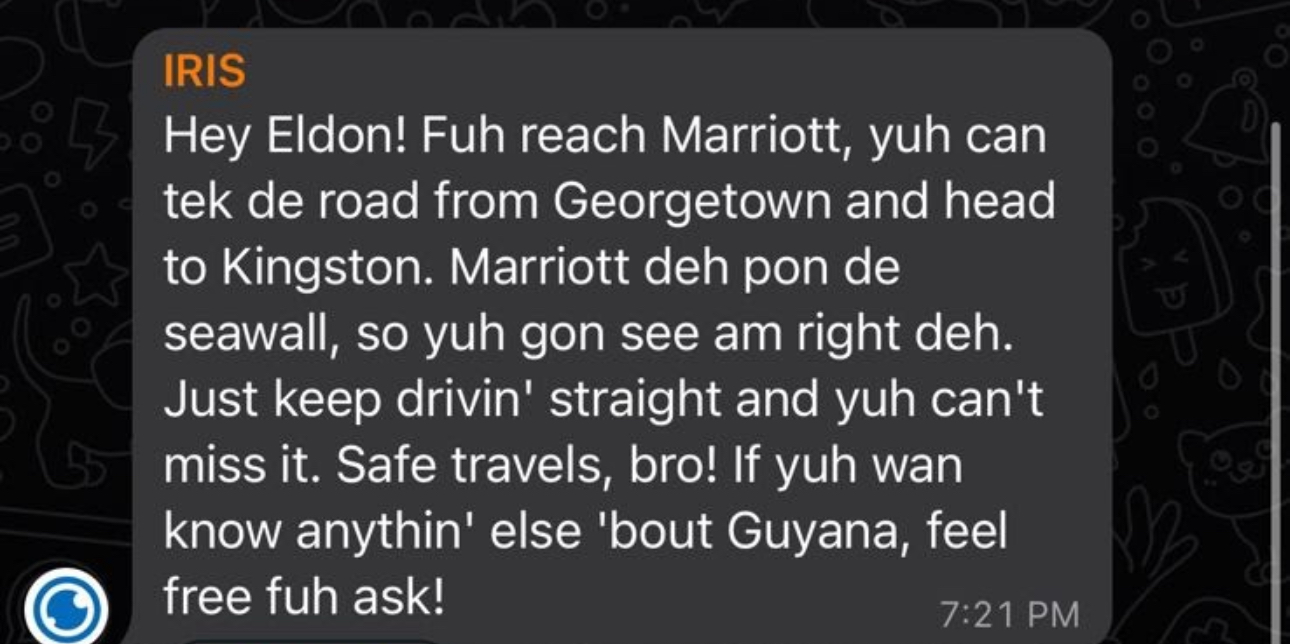} 
    \caption{Conversational Agent in Whatsapp speaking in Guyanese Creole.}
    \label{fig:iris}
    \captionsetup{font=small}
\end{figure}

One of the major issues affecting the formal adoption is Creolese despite its prominence as a spoken language is its lack of use in formal communication outlets such as literature, news, and written texts. AI-driven applications fueled by rich data sources such as \datasetname{} present a major opportunity for enabling the development of educational content, legal documents, and official communications in Creolese. Figure \ref{fig:iris} showcases a conversational AI Assistant named IRIS \footnote{\url{https://fb.watch/rbtO5Wocny/}} deployed to citizens of Guyana speaking in Creolese using \datasetname{}. Such applications present the ability to make Creolese more accessible and applicable in various formal contexts further allowing citizens to feel empowered and foster a sense of national pride.

\section{Related Works}
In the context of linguistic diversity, prior works \cite{hershcovich2022challenges, lent-etal-2021-language, lent-etal-2022-ancestor} have highlighted the challenges faced by lesser-known languages, emphasizing the importance of recognition and preservation. Works such as \citet{dabre2022kreolmorisienmt}, \citet{hagemeijer2014gulf}, and \citet{liu2022singlish} have contributed to advancing NLP research in Creole languages by building a corpus of text for various Creole languages, fostering machine translation, and enhancing language modeling techniques specific to these linguistic varieties. Our work falls into this category. On the other hand, works such as \citet{lent2022ancestor} and \citet{lent-etal-2022-creole} emphasize the importance of linguistic diversity by documenting the challenges and exploring the complexities of language modeling for underrepresented languages. The juxtaposition of these studies with the dominance of major languages in NLP underscores the need for more inclusive research efforts that consider the linguistic richness and cultural significance of smaller, indigenous languages within global technological advancements.

\section{Conclusion}
In this paper, we introduce \datasetname, a corpus of Guyanese Creolese designed to facilitate advancements in NLP research. We discuss the process of gathering and digitizing this diverse corpus while highlighting the unique opportunities presented by recent NLP advancements for accelerating the formal adoption of Creole languages in the Caribbean. By providing access to a rich collection of colloquial language expressions, idioms, and regional variations, we hope to encourage further research in this field and improve the representation and understanding of Creole languages in NLP. 

\section{Limitations}
While our work aims to contribute to the advancement of NLP for Creole, several limitations arise:

\textbf{Limited Representation}: Guyana is home to many languages outside of Creolese such as Wapichan, Makushi, Wai Wai, Akawaio, Arekuna, Patamuna, Kalina (Carib), Warrau, and Lokono to name a few. Given the cultural significance of these languages, future research should prioritize their inclusion to ensure a more inclusive and representative dataset. Additionally, The rich tapestry of languages in the region extends beyond Guyanese Creole, and efforts should be made to include additional Creole languages and dialects for a more comprehensive understanding.

\textbf{Limited Generalizability}: The findings and insights gained from our work, particularly regarding the formal adoption of Creole languages, may have limited generalizability to other regions or linguistic contexts. 

\textbf{Language Evolution}: Creole languages, by their nature, are dynamic and subject to continuous evolution. The static nature of a curated corpus and machine translation models may not fully capture the evolving linguistic landscape, necessitating regular updates and adaptations to reflect current linguistic usage.

\section*{Acknowledgements}
We thank our anonymous reviewers for their feedback and suggestions. This work is supported in part by the North American Chapter of the Association for Computational Linguistics (NAACL) Regional Americas Fund. \footnote{\url{https://www.aclweb.org/portal/content/naacl-regional-americas-fund-cfp-2023}}

\bibliography{custom}

\appendix

\section{Guyanese Creole Translation Tool}
In this section, we further showcase the Guyanese Creole Translation tool detailing our prompts and user interface.

\begin{figure}[!htb]
    \begin{verbatim}
    Translate the following Guyanese Creole text 
    and provide the resulting English translation.
    Please ensure that the translation is clear and 
    accurate. Guyanese Creole is spoken in Guyana
    and may include unique vocabulary and grammar. 
    Try to capture the original meaning while 
    making it comprehensible in English.

    Glossary:
      English: Swallow
      Creole: Swalla

      English: Stagger
      Creole: 'Taggah

      English: Stop-off
      Creole: "Taff-aff
      ...
    Translations
      Translation 1: The beef cooked until it was soft
      Text 1: Di biif kuk kuk kuk til ii saaf

      Translation 2: But my grandfather had a boat
      Text 2: Bo mi granfaada bin ga wan boot
    \end{verbatim}
    \caption{Example GPT-4 Prompt with translation examples from \citet{peirs1902proverbs}.}

\end{figure}






\end{document}